\documentclass[letterpaper, 10 pt, conference]{ieeeconf}  

\IEEEoverridecommandlockouts                              

\usepackage{graphics} 
\usepackage{epsfig} 
\usepackage{amsmath} 
\usepackage{amssymb}  
\usepackage{bm} 

\usepackage[flushmargin,hang]{footmisc} 

\usepackage{flushend} 
\usepackage{comment}	

\usepackage{siunitx} 

\usepackage{pgfplots}
\usepackage{algorithm}
\usepackage{amsmath,amssymb,amsfonts}
\usepackage{graphicx}
\usepackage{algorithm}
\usepackage{algpseudocode}
\usepackage{textcomp}
\usepackage{ifthen}
\usepackage{subcaption}
\usepackage{multirow}
\usepackage{url}
\usepackage[hidelinks]{hyperref}
\pgfplotsset{compat=newest}
\usetikzlibrary{plotmarks}
\usetikzlibrary{arrows.meta}
\usepgfplotslibrary{patchplots}
\usepackage{booktabs} 
\usepackage{grffile}
\pgfplotsset{plot coordinates/math parser=false}
\newlength\fwidth
\newlength\fheight

\usepackage{cite}

\title{\LARGE \bf
Transferable Deep Reinforcement Learning for Cross-Domain Navigation: from Farmland to the Moon} 

\author{Shreya Santra$^{*}$, Thomas Robbins$^{*}$ and Kazuya Yoshida
\thanks{
$^{*}$This work was supported by Kisuitech, Japan.
    }%
    \thanks{
    All authors are with the Space Robotics Lab. (SRL) in Department of Aerospace Engineering, Graduate School of Engineering, Tohoku University, Sendai 980--8579, Japan.  
    }%
    \thanks{
    \textit{The corresponding author is Shreya Santra. }
    }
\thanks{~~~~(E-mail: \tt{shreya.santra@tohoku.ac.jp})}
    }%

\begin{document}

\maketitle
\thispagestyle{empty}
\pagestyle{empty}


\begin{abstract}
Autonomous navigation in unstructured environments is essential for field and planetary robotics, where robots must efficiently reach goals while avoiding obstacles under uncertain conditions. Conventional algorithmic approaches often require extensive environment-specific tuning, limiting scalability to new domains. Deep Reinforcement Learning (DRL) provides a data-driven alternative, allowing robots to acquire navigation strategies through direct interactions with their environment. This work investigates the feasibility of DRL policy generalization across visually and topographically distinct simulated domains, where policies are trained in terrestrial settings and validated in a zero-shot manner in extraterrestrial environments. A 3D simulation of an agricultural rover is developed and trained using Proximal Policy Optimization (PPO) to achieve goal-directed navigation and obstacle avoidance in farmland settings. The learned policy is then evaluated in a lunar-like simulated environment to assess transfer performance. The results indicate that policies trained under terrestrial conditions retain a high level of effectiveness, achieving close to 50\% success in lunar simulations without the need for additional training and fine-tuning. This underscores the potential of cross-domain DRL-based policy transfer as a promising approach to developing adaptable and efficient autonomous navigation for future planetary exploration missions, with the added benefit of minimizing retraining costs.

\end{abstract} 


\section{Introduction}\label{introduction}
Despite decades of planetary space missions, autonomous navigation remains a critical challenge, as navigation failures can lead to mission loss. While limited teleoperation is feasible for lunar missions due to the relatively short communication delay of \SI{1.3}{s}, missions to more distant bodies such as Mars face significant latency \SI{21}{minutes} \cite{BURNS2019195}, \cite{ZHANG_Zhurong} and potential blackouts during solar conjunctions. These constraints highlight the need for reliable autonomous navigation systems that can utilize both global and local information to operate independently of Earth-based control, as demonstrated in \cite{SANTRA_2024}. Such systems may employ inertial, celestial, and vision-based navigation techniques, as in the case of the Perseverance rover \cite{Perseverace}. However, validating these algorithms requires  testing in environments that replicate the hazard characteristics of planetary surfaces, including terrain irregularities and obstacle distributions. Interestingly, recent advances in agricultural robotics present analogous challenges, with autonomous rovers navigating unstructured terrains such as orchards or fields while avoiding obstacles and following predefined paths. This convergence of requirements suggests that agricultural domains may serve as effective terrestrial analogs for navigation research on planetary surfaces. Moreover, insights gained from space robotics can reciprocally advance agricultural automation. This study investigates the bidirectional relationship, evaluating the potential for zero-shot policy transfer between these domains. The key contributions of this work are as follows:
\begin{itemize}
    \item DRL-based navigation and obstacle avoidance policy for an agricultural rover on rough terrain.
    \item Zero-shot policy transfer strategies to adapt terrestrially trained policies for extraterrestrial navigation.
    \item Cross-domain evaluation to demonstrate partial policy effectiveness without retraining.
\end{itemize}

\begin{figure}[t]
    \centering
    \includegraphics[width=\linewidth]{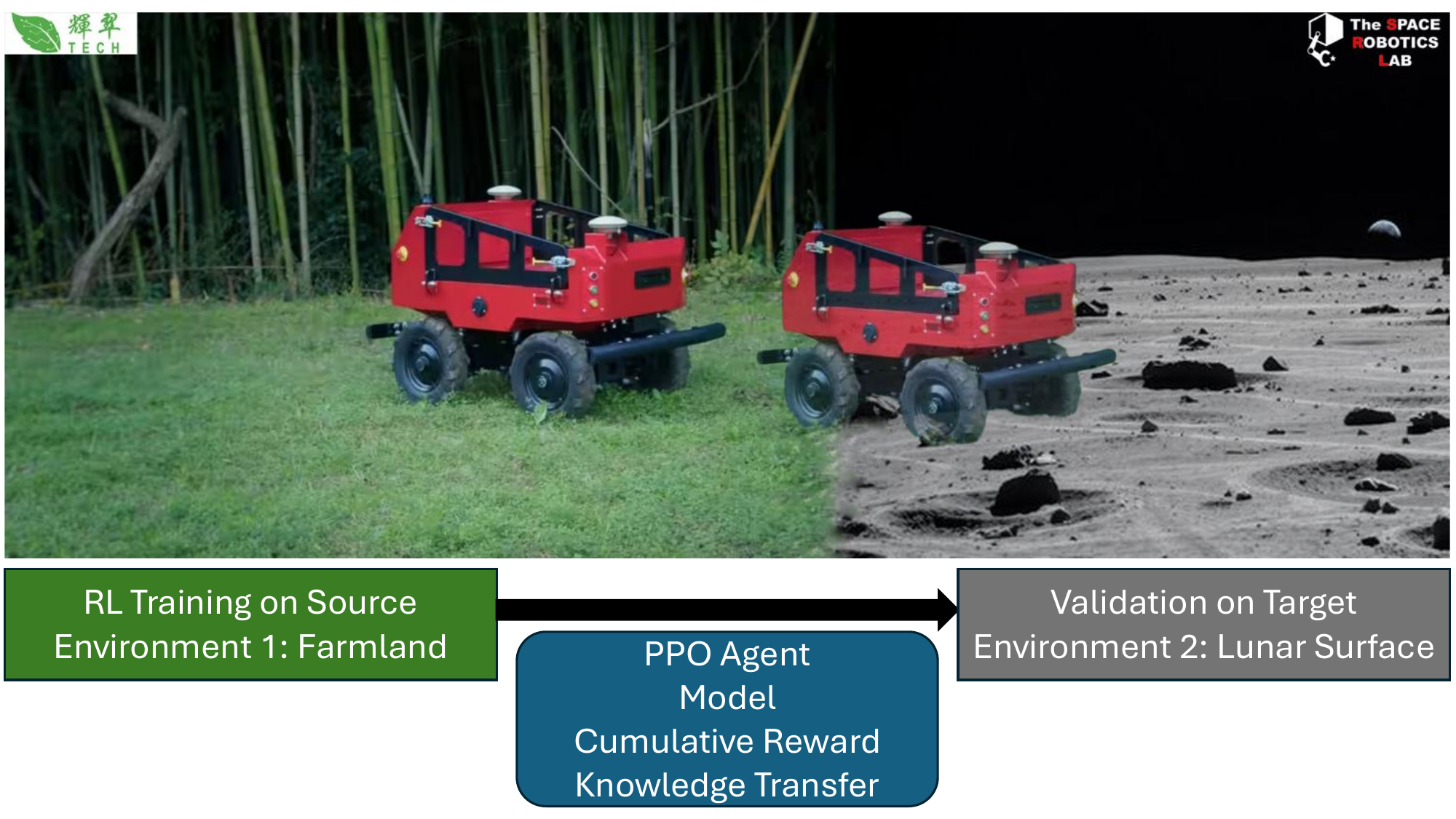}
    \caption{Transfer Learning Scenario}
    \label{fig:farm_to_moon}
\end{figure}
\section{Related Work}\label{relatedwork}

In recent years, there has been significant progress in artificial intelligence (AI)-based methods, particularly Reinforcement Learning (RL) and its deep variant DRL. By defining the task within a standard RL framework—comprising an agent, action space, observations, and reward structure—robust policies can be learned without explicit rule-based programming \cite{Sutton_RL}. RL methods are generally categorized into value-based approaches, which optimize value functions, and policy-based approaches, which directly optimize policies \cite{Byeon2023}. DRL leverages deep neural networks (DNNs) with the central motivation of enabling agents to learn behaviors interactively, optimizing actions via sensor feedback and rewards, leading to more robust and generalized policies \cite{RL_driving_Gao}. This helps address limitations inherent in teleoperation and traditional algorithmic approaches, which often rely on human intuition and may overlook critical edge cases or fail to generalize across diverse scenarios. Although DRL in robotics is a rapidly growing field with notable recent advancements \cite{RL_robotics}, its application to agricultural rover navigation and obstacle avoidance in farmland settings remains limited. Prior studies \cite{YU2023107546, Yang_Agri_RL, LIN2021106350} trained agricultural robots to enable collision-free planning in navigation and harvesting using different RL approaches. However, none of these consider competence or knowledge transfer to another field such as space exploration. Autonomous navigation and obstacle avoidance are critical for rovers in lunar, planetary, and agricultural environments with unstructured, obstacle-rich terrain. This research aims to develop adaptive navigation and obstacle avoidance for farm-assistive robots to enhance agricultural labor efficiency, while demonstrating transfer learning to space exploration by exploiting similarities between terrestrial and extraterrestrial terrains. 

Transfer learning (TL) is a machine learning technique that reuses knowledge acquired in one domain (source) to improve learning in a related domain (target). It has emerged as a means to address key challenges in diverse domains by leveraging external knowledge to improve both the efficiency and effectiveness of the learning process \cite{Fatima_TL_RL}. TL in reinforcement learning focuses on developing methods to leverage knowledge from source tasks to improve learning in a target task. When task similarities exist, this knowledge can accelerate learning and reduce the number of samples needed to achieve near-optimal performance. Research in this area aims to formalize the transfer problem and define key settings \cite{Lazaric2012}. It accelerates model adaptation and enhances performance, particularly when data are limited. 
TL approaches can be categorized as: (i) Zero-shot transfer – directly applicable to the target domain without additional training; (ii) Few-shot transfer – requiring only limited samples from the target domain; (iii) Sample-efficient transfer – improving data efficiency relative to standard RL.   In \cite{JMLR_Taylor, TL_RL, KNOWLEDGE_TRANSFER} a systematic review of TL approaches for RL is presented, examining key aspects such as the differences between source and target domains, the type of knowledge being transferred, the information available in the target domain, and the methods used for transfer. Information transferred between source and target tasks can range from low-level task-specific data, such as expected outcomes of actions in particular states, to high-level heuristics that guide learning. The effectiveness of transfer depends on task similarity, the type of knowledge transferred, the task mappings defined, and the assumptions made about the domains \cite{TL_RL}. We propose a zero-shot policy transfer framework leveraging pre-trained policies learned in a source domain (farmlands) and directly applying them to a target domain (lunar terrains), as illustrated in Fig.~\ref{fig:farm_to_moon}.



\section{Methodology}\label{methodology}
This section describes the Problem formulation, followed by the hardware and software framework.

\subsection{Problem Formulation}\label{problem}
In practice, RL trains an agent through trial and error, while its deep variant uses neural networks to handle more complex tasks and environments. An agent $A$ interacts with an environment $E$ by executing a sequence of actions $(a_o, a_1, ..., a_i)$, which, in turn, alter the state of $E$. The resulting changes are perceived through observations $(b_0, b_1,..., b_j)$,  and evaluated using reward signals $(r_0, r_1, ..., r_k)$.
The goal is to learn a policy that maximizes cumulative reward over action–observation cycles, making reward design and hyperparameter tuning essential. In this study, a four-wheeled rover is the agent, with actions as four wheel speeds and observations including goal distance, position, and velocity. Rewards increase with proximity to the goal and collision avoidance, encouraging safe, efficient navigation. The RL problem is formulated as a Markov Decision Process (MDP) and solved using suitable optimization strategies\cite{RLrewards_Diverse} such as $\mathcal{(S, A, P, R, \gamma)}$ where:
\begin{itemize}
    \item $\mathcal{S} = {s_0, s_1, ..., s_i}, i \in N^*$ describes the possible states, which represent environmental configurations.
    \item $\mathcal{A} = {a_0, a_1, ..., a_i}, i \in N^*$ describes the actions that the agent can take.
    \item $\mathcal{P}(s_{t+1}|s_t, a_t), (s_t, s_{t+1})  \in \mathcal{S^2}, a_t \in \mathcal{A}$ describes the probability for the agent to transition to state $s' = s_{t+1}$ from state $s_t$ as a result of action $a_t$. As a probability function, it fulfill
    \begin{equation}
    \sum_{s^{\prime} \in \mathcal{S}} \mathcal{P}\left(s^{\prime} \mid s, a\right)=1 \forall(s, a) \in \mathcal{S} \times \mathcal{A}
    \end{equation}
    \item $\mathcal{R}(s, a)$ is the reward function that awards values for taking action $a$ in state $s$.
    \item $\mathcal{\gamma} \in [0;1]$ is a discount factor, responsible for the importance of short-term or long-term rewards.
\end{itemize}
    
These points are factored into a policy $\pi(a|s)$, a probability distribution that
models the agent’s current behavior, i.e., maps states to actions. Therefore, optimization of the agent’s behavior is achieved through an optimized policy $\mathcal{\pi^*}$ that maximizes the expected return:

\begin{equation}
    \begin{array}{l}
\pi^{*} =\arg \max _{\pi} \mathbb{E}_{\pi}\left(\sum_{k=0}^{\infty} \gamma^{t} r_{t+k}\right)\\
\end{array}
\end{equation}
 where $(s_{t}, a_{t}) \in \mathcal{S} \times \mathcal{A}\forall{t}$ are states and actions at time $t$.


The Proximal Policy Optimization (PPO), introduced by Schulman et. al. \cite{PPO1Schulman2017PPO} limits extreme policy updates from large probability shifts by clipping gradients with an adaptive Kullback-Leibler penalty, improving stability and allowing multiple training epochs for safer optimization outlined in Eq.~\ref{PPOclip} and ~\ref{PPO_KL}.  

\begin{equation}
    L_{CLIP}(\theta) = \hat{\mathbb{E}}_t \left[ \min\left( r_t(\theta)\hat{A}_t, \text{clip}(r_t(\theta), 1 - \epsilon, 1 + \epsilon) \hat{A}_t \right) \right]
    \label{PPOclip}
\end{equation}

\begin{equation}
    L_{KLPEN}(\theta) = \hat{\mathbb{E}}_t \left[ r_t(\theta) \hat{A}_t - \beta \text{KL}[\pi_{\theta_{\text{old}}}(\cdot | s_t), \pi_\theta(\cdot | s_t)] \right]
    \label{PPO_KL}
\end{equation}
where, $KL$ divergence, measures how one probability distribution differs from another.

\subsection{Hardware Configuration}\label{hardware}
\begin{figure}[t]
    \centering
    \includegraphics[width=\linewidth]{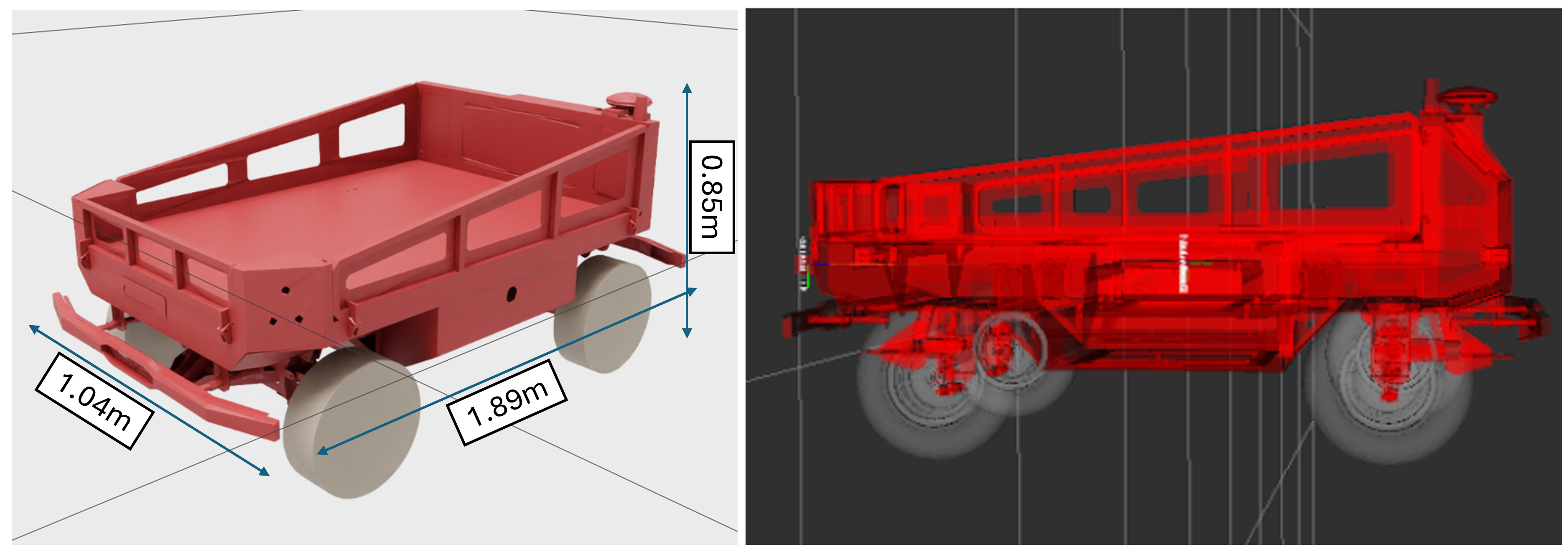}
    \caption{3D model and URDF of Adam Rover}
    \label{fig:adam_rover}
\end{figure}

The agent selected for this study is KisuiTech’s Adam rover, depicted in Fig.~\ref{fig:adam_rover}. Inspired by lunar rovers, Adam rover navigates rugged farm terrain just like it would on the Moon. Developed to support agricultural operations, KisuiTech aims to enhance cost-performance efficiency by automating key tasks in the harvesting process \cite{kisui}. The Adam rover assists farmers by transporting harvested produce from the field to central storage areas, such as trailers positioned at the orchard perimeter. Acting as an intermediate carrier, it collects produce directly from workers, minimizing the need for repeated manual trips using baskets or small carts. This reduces physical strain on workers, streamlines logistics, and contributes to faster and more efficient harvesting. The rover model is $z=1.89m$ long, $x=1.12m$ wide and $y=0.77m$ tall. The Adam rover uses a four-wheel differential drive configuration with a steering behavior similar to skid-steer motion. The wheel speed equations for a four-wheel differential drive are as follows:
\begin{equation}
    \text { Robot's Linear Velocity (v): } v=(v r+v l) / 2 \text {, }
\end{equation}
where $vr$ is the right wheel's linear velocity and $vl$ is the left wheel's linear velocity.
\begin{equation}
    \text { Robot's Angular Velocity }(\omega): \omega=(v r-v l) / L,
\end{equation}
where $L$ is the distance between the wheels.
From a kinematic point of view, Adam rover fulfills the requirements to be a non-holonomic 4-wheel skid-steering differential drive system as in space exploration rovers. To enable obstacle detection and avoidance, rover is equipped with front and rear RGB-D cameras. 

\subsection{Simulation Framework} \label{simulation}
This project uses NVIDIA Isaac Sim 4.2 \cite{isaac21} with its DRL-focused Isaac Lab 1.0 \cite{isaac22}, both built on NVIDIA Omniverse for high-fidelity physics and realistic robotics environments on a RTX4080 GPU. The Adam Rover was modeled in SolidWorks, exported to URDF for ROS 2 compatibility, and simplified for simulation efficiency. Imported into Isaac Sim, the URDF is converted to Universal Scene Description (USD), where the rover’s articulated root body connects to wheels via ideal revolute joints for physics simulation. Isaac Lab accesses key states such as position and orientation, which—after coordinate corrections—are sufficient to compute the rover’s distance to the goal.
To better approximate real-world conditions, a custom terrain is integrated into the environments using the \textit{HfRandomUniformTerrainCfg} class from the Isaac Lab framework~\cite{isaac22}. The parameters governing this terrain generation are detailed in Tables~\ref{tab:terrain-gen-baseline} and \ref{tab:sub-terrain-params}. Two parameter sets are provided to the terrain generator: one defines the main three-dimensional (3D) terrain scales, and the other specifies noise patterns for smaller sub-terrain squares that compose the larger surface. This noise adds realism to rover navigation environments. Although the two sub-terrain models used here are identical, the dual-terrain setup was retained for testing with varied parameter distributions and for future flexibility. Noise amplitudes between \SI{0.03}{m} and \SI{0.07}{m} produced roughness similar to soil, offering a balance between unrealistic flatness and excessive inclines. 

\begin{table}[ht]
\centering
\caption{Terrain generator baseline values}
\label{tab:terrain-gen-baseline}
\begin{tabular}{|l|l|}
\hline
\textbf{General parameters} & \textbf{Values} \\ \hline
Vertical scale              & 0.005           \\ \hline
Horizontal scale            & 0.1             \\ \hline
Slope threshold             & 0.75            \\ \hline
\end{tabular}

\end{table}

\begin{table}[ht]
\centering
\caption{Sub-terrain parameters}
\begin{tabular}{|l|l|l|}
\hline
Sub-terrain parameters & Sub-terrain 1    & Sub-terrain 2    \\ \hline
Proportion             & 0.5              & 0.5              \\ \hline
Noise range            & {[}0.03; 0.07{]} & {[}0.03; 0.07{]} \\ \hline
Noise step             & 0.01             & 0.25             \\ \hline
Border width           & 0.01             & 0.25             \\ \hline
\end{tabular}

\label{tab:sub-terrain-params}
\end{table}

%

These parameters control the generation of randomized patterns across the terrain, where the generator assigns a sub-terrain texture to each segment—defined by row and column indices—by sampling from the available sub-terrain indices. 

\subsection{Environment Setup}\label{simulations}

\begin{figure}[t]
    \centering
    \includegraphics[width=\linewidth]{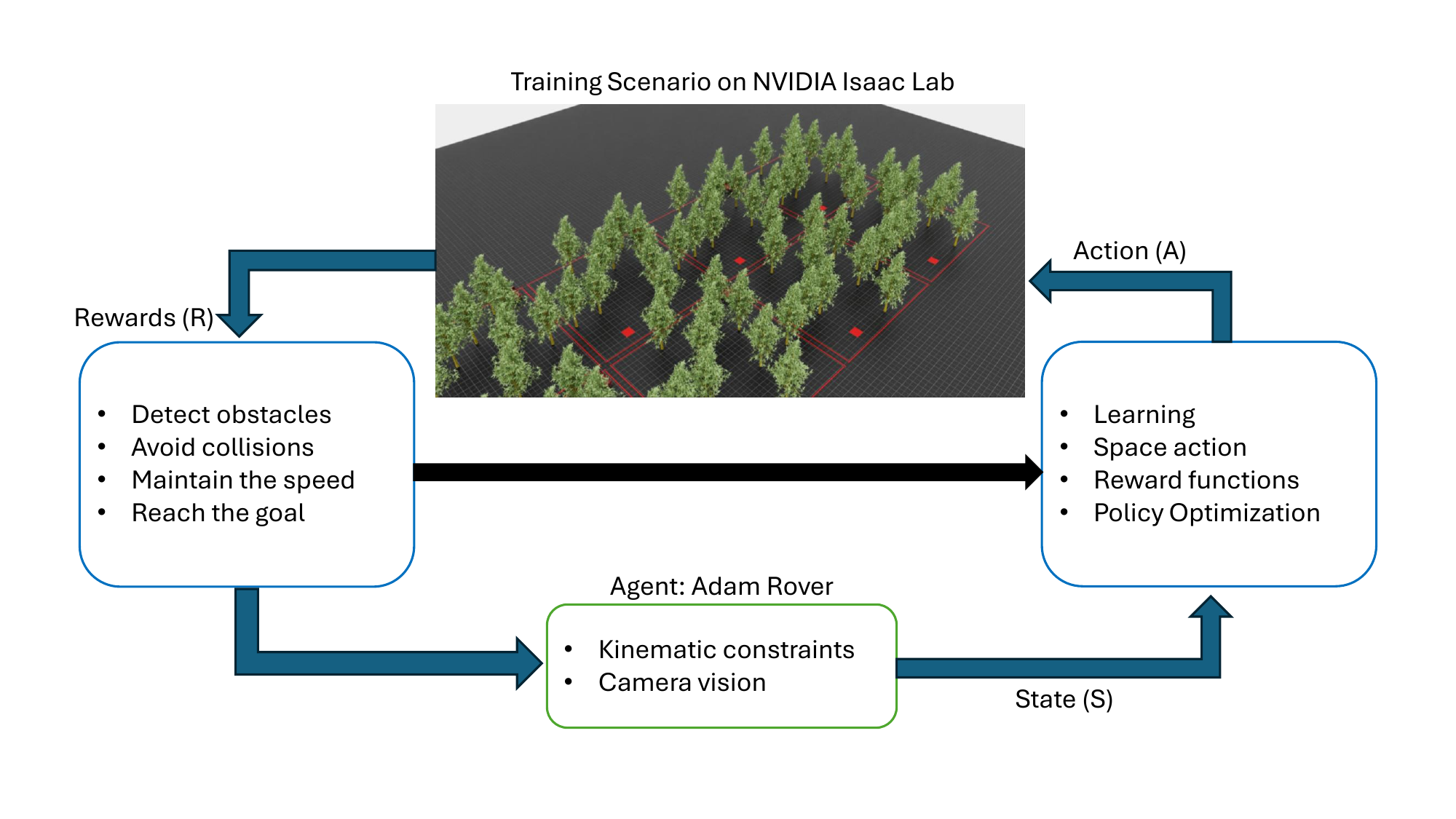}
    \caption{An overview of the DRL framework for training in multiple parallel environments}
    \label{fig:train-environment}
\end{figure}

\begin{figure}[t]
    \centering
    \includegraphics[width=\linewidth]{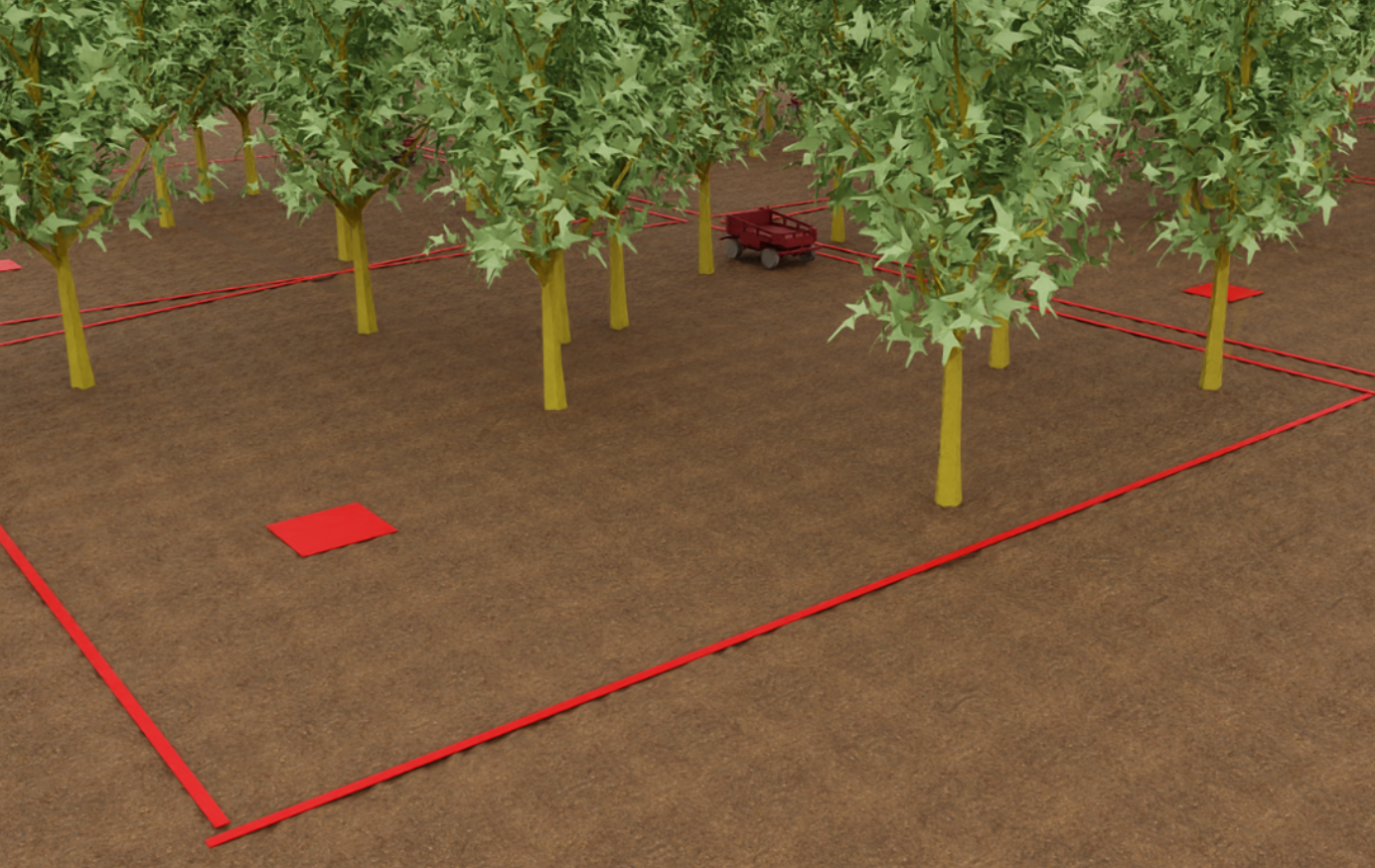}
    \caption{Adam Rover navigating on rough terrain in farm-like environment}
    \label{fig:agri-environment}
\end{figure}

\begin{figure}[t]
    \centering
    \includegraphics[width=\linewidth]{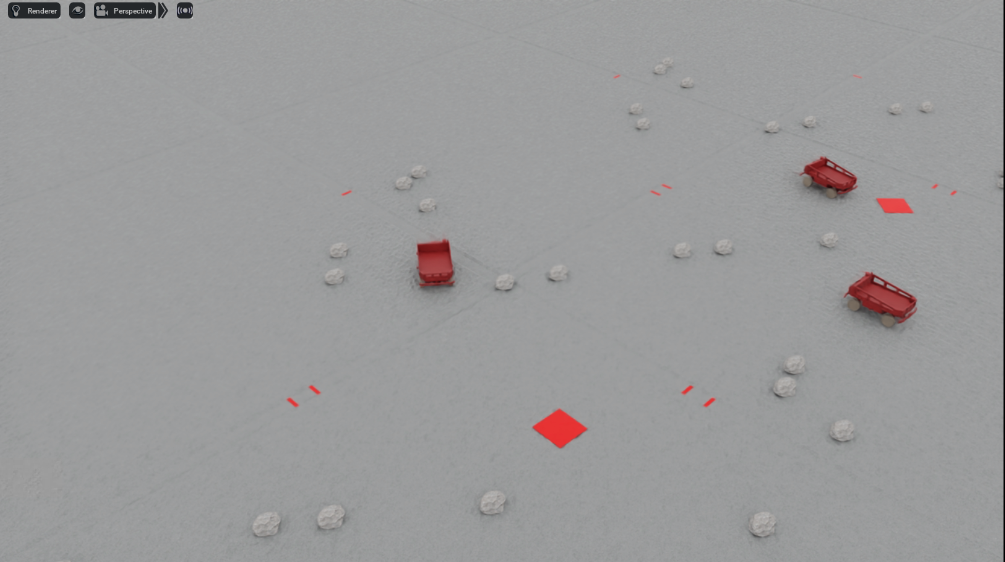}
    \caption{Validation on Simulated Lunar Terrain}
    \label{fig:lunar-environment}
\end{figure}

This section explains the design of the simulated environments to evaluate safe rover navigation under controlled conditions. The tasks simulate scenarios such as an agricultural rover returning to storage or a lunar rover reaching a target site. Each environment defines a start point, goal, boundaries, and obstacles to support learning. Multiple parallel environments are run in Isaac Sim and Isaac Lab to accelerate training, limited by computational resources as shown in Fig.~\ref{fig:train-environment}. Implemented as a MDP, the Adam rover is the learning agent, and the environment dimensions are scaled to fit its size without excessive constraints. A \SI{15}{m} by \SI{15}{m} square, which is a realistic local operational area in an orchard and for planetary exploration, was considered for the training environment. 
The start and goal positions are assigned randomly, with ten obstacles introduced through a randomized distribution. To reflect features typical of the agricultural environments underlying the training setup, the obstacles are modeled as trees. In subsequent validation simulations representing lunar terrains (see Fig.~\ref{fig:lunar-environment}), these were replaced with boulders. Fig.~\ref{fig:agri-environment} illustrates a representative instance of a single environment on field, with the start location marked in the lower-left corner and the goal location indicated by the red square situated on the opposing side, while obstacle positions are denoted by green trees.

Simulation parallelization was limited by computational load, driven mainly by terrain complexity and dual-camera data per rover. Tests showed that sixteen environments exceeded practical limits, even at reduced camera resolution. A configuration of eight environments, though slower, ensured stable performance without degrading simulation quality. Episode length, defining the time allowed for task completion, was tuned to balance learning efficiency: short episodes risk premature failure, while overly long ones dilute reward signals. A default length of 125,000 timesteps (2–3 hours real time) was used, with some runs shortened to ~100,000 timesteps when learning was sufficient to determine task success or failure.

\subsubsection{Observations and Rewards}
Observations are selected input variables critical to the learning process, enabling the model to infer which configurations yield higher rewards. For optimizing the Adam rover’s navigation and collision avoidance, the size of the observation space is 12, that includes:
\begin{itemize}
    \item The rover’s 2D pose in the environment’s reference frame $(x_A, y_A)$.
    \item Relative distance to the goal along the x and y $(d_x, d_y)$.
    \item Orientation to the goal $(\theta)$ and heading angle with respect to the x-axis $(\beta)$.
    \item Front and rear avoidance filters $f_{front}$ and $f_{rear}$ help the model associate their activation with the presence of obstacles.
    \item Four wheel speeds $(v_{W1}, v_{W2}, v_{W3}, v_{W4})$ allow the model to infer the direction of movement.
\end{itemize}

The allowable actions for the Adam rover are constrained to motion control, effectively reducing the action space to two degrees of freedom, namely the linear and angular speeds of the main body, $A = v_{body}, \omega_{body}$. Resets define episode outcomes and ensure simulation continuity. An episode succeeds if the rover reaches the goal and fails if it goes out of bounds, times out, or collides with an obstacle. In all cases, the rover is reset to its initial state to start a new episode. Resets at the end of an episode are crucial for optimization, teaching the rover to distinguish success from failure. 
The PPO implementation follows the clip objective (Eq. 3). Training used $n-steps$ = 2048, $batch-size$ = 16384, $n-envs$ = 8, $total-env-steps = 1.2 × 10^7$, network = [128, 128, 64], and seeds = {0,…,4}.
The rewards and penalties considered for the training are listed in Table~\ref{table:rewards_penalty}. These values are instrumental in iteratively determining what is prioritized during the optimization process. Distance and orientation rewards use hyperbolic tangent functions shown in Eq. ~\ref{equation_rewards} instead of linear scaling, as linear rewards failed to prevent incorrect rotations, backtracking, or random movements in earlier navigation tests.
At the start of an episode, the rover receives small positive rewards primarily from $r_{alive}$ and $r_{angle}$ encouraging fine-tuned behaviors such as adjusting wheel speeds and initiating correct movement toward the goal. As the simulation progresses, these minor rewards become less influential, allowing for adaptive behaviors like detours or reversing to avoid obstacles and penalties before resuming optimal navigation.

\begin{equation}
    \begin{array}{c}
r_{\text {distance}}(d)=3\left(0.8-\tanh \left(\frac{d}{15}\right)\right) \\
\\
r_{\text {angle}}(\theta)=0.6\left(-0.9-\tanh \left(\frac{\theta-1}{0.1}\right)\right)
\end{array}
\label{equation_rewards}
\end{equation}
\begin{table}[ht]
\caption{Rewards Table}
\label{table:rewards_penalty}
    \centering
    \begin{tabular}{|c|c|c|} 
        \toprule
        \textbf{Reward or Penalty} & \textbf{Values} & \textbf{Significance}\\
               \midrule
        $r_{alive}$ & +1 & Given when\\
        &  & rover is not terminated\\\hline
        $r_{distance}(d)$ & see Eq. ~\ref{equation_rewards} &  Increases based on the rover’s  \\
        &  & closeness to the goal\\\hline
        $r_{angle}(\theta)$ & see Eq. ~\ref{equation_rewards} & Increases based on the rover’s  \\
        &  & orientation relative to the goal\\\hline
        $r_{success}$ & +20  & A large reward given upon  \\
        &  & successfully reaching the goal\\\hline
        $r_{reset}$ &  & When the rover resets after a \\
        $p_{reset}$ &  $\pm 2$ &  successful or failed episode\\\hline
        $r_{forward}$ & +0.6  &  When the wheels move  \\
        &   &  forward towards the goal \\\hline
        $r_{backward}$ &  & Wheels move
        backward from \\
        $p_{backward}$ &  $\pm 0.3$ & goal when obstacle is detected\\\hline
        $r_{speed}$ & $\pm 0.6$  &  Wheel speeds are below or\\ 
        $p_{speed}$ &   & above the instability threshold \\\hline
        $p_{obtacle}$ & -1  & Obstacle is detected \\\hline
    \end{tabular}
\end{table}

\subsubsection{Hyperparameter Tuning}
The hyperparameters for the Stable Baseline3's\cite{stable-baselines3, PPO2OpenAIbaselines} PPO algorithm used in this work are listed in Table ~\ref{table:hyperparameters}.

\begin{table}[ht]
\caption{Hyperparameters}
\label{table:hyperparameters}
    \centering
    \begin{tabular}{|c|c|} 
        \toprule
        \textbf{Hyperparameter} & \textbf{Values} \\
               \midrule
        Discount Factor $(\gamma)$ & 0.99 \\\hline
        Generalized Advantage Estimate (GAE) $(\lambda)$ & 0.95 \\\hline
        Learning Rate & $5 \times 10^{-4}$   \\\hline
        KL Threshold ($\beta$) & 0.008   \\\hline
        Max Epochs & 150 \\\hline
        Clipping Parameters ($e_{clip}$) & 0.2 \\\hline
        Minibatch Size & 16384 \\\hline
        Mini Epochs & 8 \\\hline
        Critic Coefficient & 4 \\\hline
        Sequence Length & 4 \\\hline
        Bound Loss Coefficient & 0.0001 \\\hline
    \end{tabular}
\end{table}

\section{Learning, Training and Validation} \label{l_t_V}

Data collection was conducted using a reward tracker that computed the average reward—defined as the total reward across all rovers divided by their number—throughout the simulation and plotted it against the corresponding timesteps. Multiple runs of the same model were sometimes required to obtain a trained policy that met performance expectations. The focus of this approach was not on reward convergence but on the frequency of reward spikes, which represent successful episode completions. Each spike corresponds to a rover reaching its goal, with higher spike density indicating improved task success rates across the rover population.

\subsection{Learning on Agriculture Environment}
\begin{figure}[t]
    \centering
    \includegraphics[width=\linewidth]{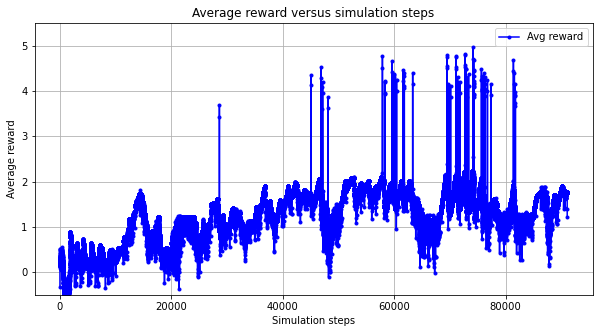}
    \caption{Average reward versus timesteps for one of the trained models in the navigation and avoidance on farmland scenario}
    \label{fig:learning_results}
\end{figure}

Trainings were conducted using the simulation environment described in the previous section. A reward tracker recorded the average reward per rover plotted against time, with training progress assessed by the frequency of reward spikes—indicating successful episodes—rather than overall reward convergence. Each spike corresponds to a rover reaching its goal, with higher spike density indicating improved task success rates across the rover population shown in Fig.~\ref{fig:learning_results}. “Reaching the goal” is defined as the rover entering within a \SI{0.5}{m} radius of the center of the target red square. The model exhibited promising yet unstable performance for navigation and obstacle avoidance, characterized by alternating phases of success and reset loops. Initial instability subsided after 10,000 timesteps, followed by increasing rewards and intermittent reset loops, which occur when the rover infers an "optimal" path that coincides with an obstacle, resulting in repeated collisions until the penalties finally encourage change. Over time, success phases became longer (up to 70,000–82,000 timesteps), indicating progressive optimization. The simulation was stopped at this stage to avoid overfitting while preserving the state of the trained model.

Isaac Sim automatically saves model checkpoints—configurations that achieve the highest rewards—both at fixed timestep intervals and at the end of training, providing multiple candidate models for subsequent validation experiments.


\subsection{Learning Transferability}
\begin{figure}[t]
    \centering
    \includegraphics[width=0.95\linewidth]{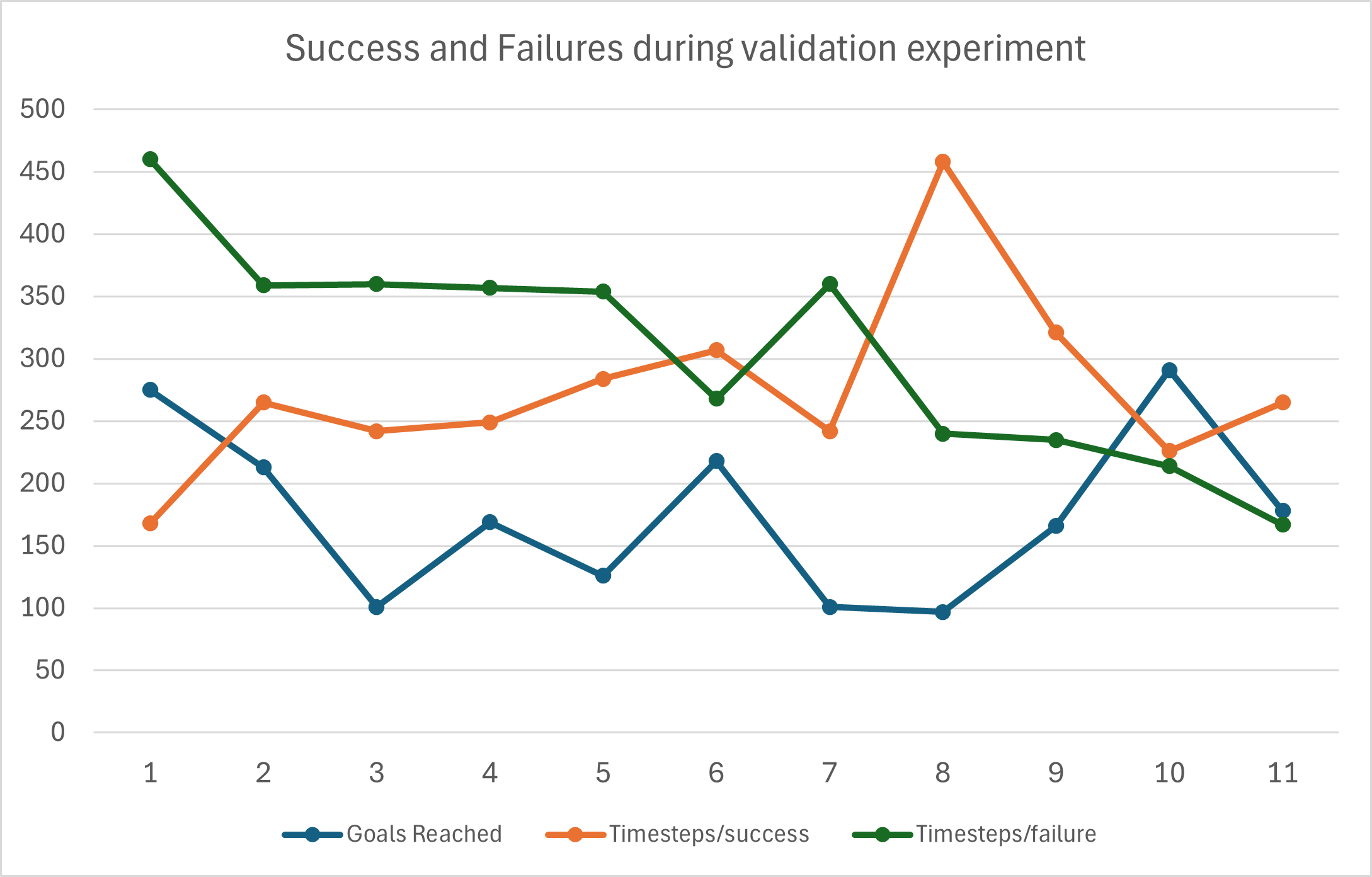}
    \caption{Success and failure performance of transferred policy}
    \label{fig:validation-plot}
\end{figure}

The aim is to show the application of a trained model in order to evaluate performance without further learning. The best performing trained model was chosen with a checkpoint around the 75000 timestep checkpoint. The validation environment was modified to evaluate the model’s ability to generalize beyond its training conditions. Tree obstacles were replaced with gray lunar rocks slightly shorter than the rovers, introducing a novel color variable to test whether obstacle detection relied on specific color–depth patterns or broader avoidance strategies. The terrain retained its three-dimensional profile generation but was assigned a lunar-gray texture visually similar to the rocks, reducing color contrast compared to the training setup. The gravity is modified to lunar gravity (\SI{1.62}{m/s^2}) and the friction coefficient to $0.45$ for lunar regolith. Obstacle positions were randomized for each validation run to assess whether the model had memorized fixed navigation paths or had learned adaptable obstacle avoidance behavior. Unlike training, where goal locations varied, a fixed goal was set at coordinates outside the previously traversed regions, ensuring that the model could not reuse its learned optimal paths. These modifications provided a controlled yet challenging environment for assessing the robustness and adaptability of the trained navigation policy.
As for zero-shot transfer simulations, which are no longer learning scenarios but practical applications of learned behavior, no fixed episode length was set. Instead, runs continued until enough successes or failures were observed (the shortest run included 169 combined outcomes). To address variations in run length, results were normalized using metrics such as timesteps per success and timesteps per failure for comparison; i.e., if the timestep-to-success ratio is smaller than the timestep-to-failure ratio, then there were more successes and vise versa, as shown in Fig.~\ref{fig:validation-plot}. The primary performance metric crucial for evaluating the effectiveness of the trained model within a given simulation is the success rate calculated as the ratio of successful episodes to the total number of attempts (successes plus failures). 
 

\section{Results and Discussions}\label{results_discussions}
A total of 10 simulation runs were conducted on the lunar environment and an average success rate of 46.69\% was observed, with the best case scenario being 73.33\%. This approach proved to be sample-efficient transfer with no pre-training on lunar data. The distributions of the obstacles and the starting point of the rover in each environment played a significant role in the success and failure of the validation experiments. In all cases, the rovers successfully avoided collisions with the obstacles. However, the time taken to generate a mean-bearing trajectory increased the number of time steps. Because collisions reset the rover immediately, obstacles located near the start cause a disproportionately higher number of failures compared to successful navigation cases.
While this work adopts terminology from transfer learning, the experiments conducted represent zero-shot cross-domain evaluation—applying a policy trained in a terrestrial domain directly to a lunar-like environment without retraining. This setting isolates the policy’s generalization capacity across visually and structurally distinct terrains rather than focusing on adaptive transfer.

\section{Conclusions}\label{conclusion}

This study provides an initial demonstration of a cross-domain test framework for evaluating DRL navigation policies across terrestrial and planetary analogs using a unified simulation pipeline. The average success rate of a zero-shot transfer of a trained policy from terrestrial to extraterrestrial settings reached nearly 50\%, despite the absence of additional learning in the new environment and the introduced changes, relying solely on the competencies acquired in the farmland setting. While the results are promising, future work should optimize model architectures, improve simulation fidelity, and conduct sim-to-real experiments to validate transferability under extraterrestrial conditions. Refining environmental design and scaling evaluations will further support reliable autonomous deployment for long-duration space missions.



\bibliographystyle{IEEEtran}
\bibliography{reference.bib}

\end{document}